\documentclass[10pt,twocolumn,letterpaper]{article}

\usepackage{cvpr}
\usepackage{times}
\usepackage{epsfig}
\usepackage{multirow}
\usepackage{graphicx}
\usepackage{amsmath}
\usepackage{amssymb}
\usepackage{lineno}
\usepackage{setspace}

\usepackage[breaklinks=true,bookmarks=false]{hyperref}

\cvprfinalcopy 

\def\cvprPaperID{****} 

\begin{document}

\title{\textbf{A Computer Vision Pipeline for Automated Determination of Cardiac Structure and Function and Detection of Disease by Two-Dimensional Echocardiography} }
\author{Jeffrey Zhang$^{1}$$^{2}$ \and
Sravani Gajjala$^{1}$ \and 
Pulkit Agrawal$^{2}$ \and 
Geoffrey H. Tison$^{3}$ \and
Laura A. Hallock$^{2}$ \and
Lauren Beussink-Nelson$^{4}$ \and
Mats H. Lassen$^{3}$ \and
Eugene Fan$^{3}$ \and
Mandar A. Aras$^{1}$$^{3}$ \and
ChaRandle Jordan$^{3}$ \and
Kirsten E. Fleischmann$^{3}$ \and
Michelle Melisko$^{3}$ \and
Atif Qasim$^{3}$ \and
Alexei Efros$^{2}$ \and
Sanjiv J. Shah$^{4}$ \and
Ruzena Bajcsy$^{2}$ \and
Rahul C. Deo$^{1}$$^{3}$$^{5}$$^{6}$$^{7}$$^\ast$\\
\\
\and
\normalsize{$^{1}$Cardiovascular Research Institute, University of California, San Francisco, USA.}\and
\normalsize{$^{2}$Department of Electrical Engineering and Computer Science, University of California at Berkeley, USA.}\and
\normalsize{$^{3}$Department of Medicine, University of California, San Francisco, USA.}\and
\normalsize{$^{4}$Division of Cardiology, Department of Medicine, and Feinberg Cardiovascular Research Institute}\and
\normalsize{Northwestern University Feinberg School of Medicine, Chicago, IL.}\and
\normalsize{$^{5}$Institute for Human Genetics, University of California, San Francisco, USA.}\and
\normalsize{$^{6}$California Institute for Quantitative Biosciences, San Francisco, USA.}\and
\normalsize{$^{7}$Institute for Computational Health Sciences, University of California, San Francisco, USA.}\and
\normalsize{$^\ast$To whom correspondence should be addressed; E-mail: rahul.deo@ucsf.edu}}


\maketitle


\pagebreak

\begin{abstract}
Automated cardiac image interpretation has the potential to transform clinical practice in multiple ways including enabling low-cost serial assessment of cardiac function by non-experts in the primary care and rural setting. We hypothesized that advances in computer vision could enable building a fully automated, scalable analysis pipeline for echocardiogram (echo) interpretation. Our approach entailed: 1) preprocessing of complete echo studies; 2) convolutional neural networks (CNN) for view identification, image segmentation, and phasing of the cardiac cycle; 3) quantification of chamber volumes and left ventricular mass; 4) particle tracking to compute longitudinal strain; and 5) targeted disease detection. CNNs accurately identified views (e.g. 99$\%$ for apical 4-chamber) and segmented individual cardiac chambers. The resulting cardiac structure measurements agreed with study report values [e.g. median absolute deviations (MAD) of 11.8 g/kg/m$^2$ for left ventricular mass index and 7.7 mL/kg/m$^2$ for left ventricular diastolic volume index, derived from 1319 and 2918 studies, respectively]. In terms of cardiac function, we computed automated ejection fraction and longitudinal strain measurements (within 2 cohorts), which agreed with commercial software-derived values [for ejection fraction, MAD=5.3$\%$, N=3101 studies; for strain, MAD=1.5$\%$ (n=197) and 1.6$\%$ (n=110)], and demonstrated applicability to serial monitoring of breast cancer patients for trastuzumab cardiotoxicity. Overall, we found that, compared to manual measurements, automated measurements had superior performance across seven internal consistency metrics (e.g. the correlation of left ventricular diastolic volumes with left atrial volumes) with an average increase in the absolute Spearman correlation coefficient of 0.05 (p=0.02). Finally, we used CNNs to develop disease detection algorithms for hypertrophic cardiomyopathy and cardiac amyloidosis, with C-statistics of 0.93 and 0.84, respectively.  Our pipeline lays the groundwork for using automated interpretation to support point-of-care handheld cardiac ultrasound and large-scale analysis of the millions of echos archived within healthcare systems.
\end{abstract}

\section{Introduction}
Cardiac muscle disease often progresses for years prior to the onset of symptoms. This process, known as cardiac remodeling, can accompany conditions such as valvular disease, hypertension and diabetes mellitus, and result in pathologic changes to the heart that are difficult to reverse once established \cite{hill}. Although early evidence of remodeling is often detectable by imaging \cite{ishizu} and could in principle be tracked longitudinally in a personalized manner, the cost of imaging all individuals with cardiac risk factors would be prohibitive. 
Automated image interpretation could enable such monitoring at far lower costs, especially when coupled with inexpensive data acquisition. For echocardiography one such strategy could involve handheld ultrasound devices used by non-experts \cite{neskovic} at point of care locations (e.g. primary care clinics) and a cloud-based automated interpretation system that assesses cardiac structure and function and compares results to one or more prior studies. Automated image interpretation could also enable surveillance of echo data collected at a given center and could be coupled with statistical models to highlight early evidence of dysfunction or detect rare myocardial diseases. Such an approach could, for example, enable systematic comparison across the tens of millions of echocardiograms completed each year in the Medicare population alone \cite{andrus}.

Automated image interpretation falls under the discipline of computer vision, which, in turn, is a branch of machine learning where computers learn to mimic human vision \cite{szeliski}.  Although the application of computer vision to medical imaging has been longstanding \cite{bajcsy}, recent advances in computer vision algorithms, processing power, and a massive increase in digital labeled data has resulted in a striking improvement in classification performance for several test cases, including retinal \cite{gulshan} and skin disease \cite{esteva}. Echocardiography, nonetheless, presents challenges beyond these examples. Rather than comprising a single still image, a typical echo study consists of up to 70 videos collected from different viewpoints, and viewpoints are not labeled in each study. Furthermore, measurements can vary from video to video because of intrinsic beat-to-beat variability in cardiac performance as well as variability from the process of approximating a three-dimensional object using two-dimensional cross-sectional images. Given the extent of this variability and the sheer amount of multidimensional information in each study that often goes unused, it appears that echocardiography would benefit from an automated learning approach to assist human interpretation.

In this manuscript, we present a fully automated computer vision pipeline for  interpretation of cardiac structure, function, and disease detection using a combination of computer vision approaches. We demonstrate the scalability of our approach by analyzing $>$4000 echo studies and validate our accuracy against commercial vendor packages. We describe some of the challenges we encountered in the process as well as potential promising applications.  We have deployed our pipeline on Amazon EC2 at http:///www.echocv.org and make source code and model weights available.

\section{Results}
\subsection{Overview: A Computer Vision Pipeline for Automated 2D-Echocardiogram Interpretation}
\begin{figure}[t!]
  \centering
  \includegraphics[width=1.0\linewidth]{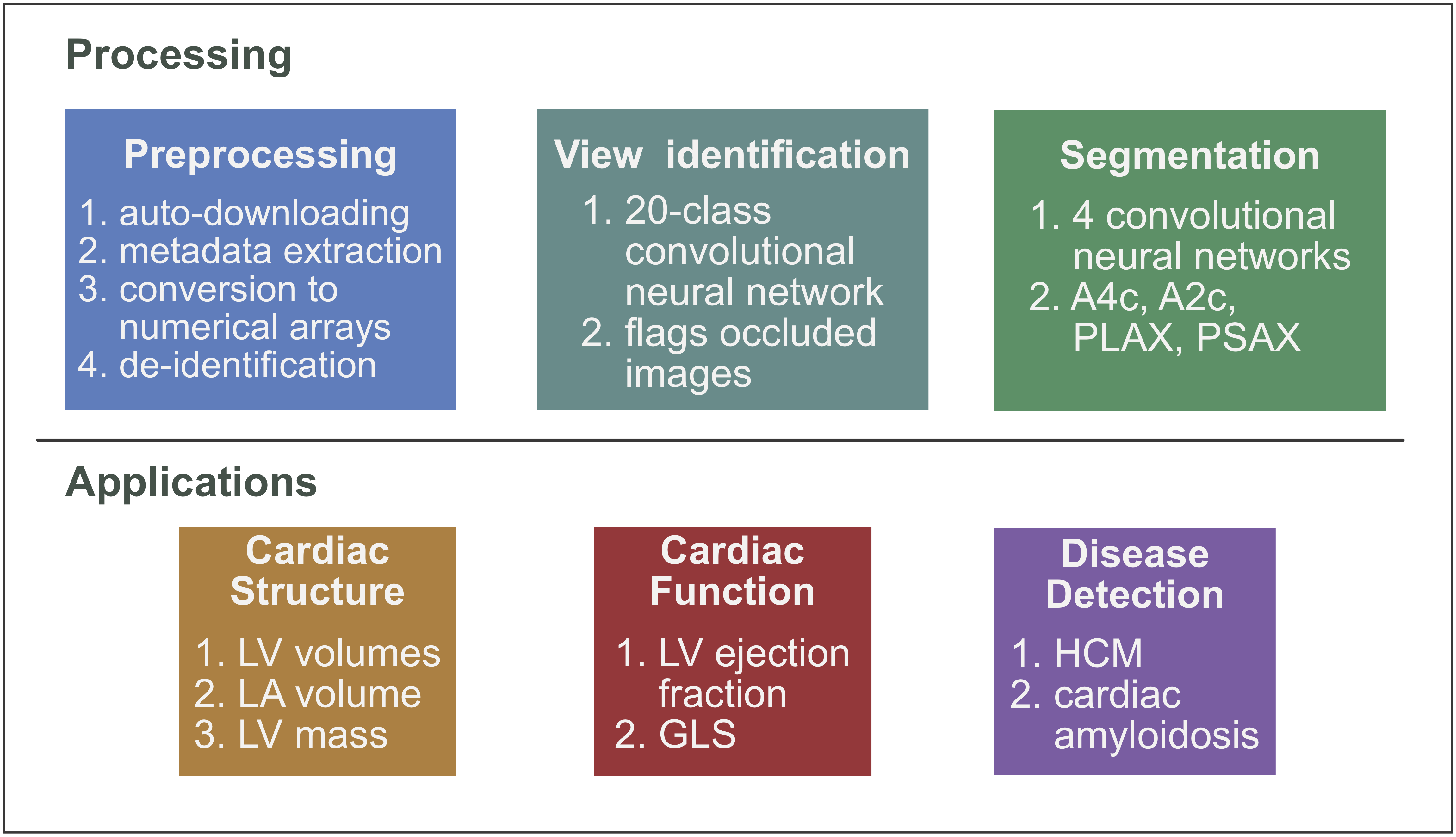}\\
  \caption[]{Overview of Automated Cardiac Interpretation Pipeline. We divided the pipeline into 6 steps. Preprocessing involves automated downloading of echo studies in DICOM format, separating video loops from still images, extracting metadata, converting them into numerical arrays for matrix computation and de-identifying images by overwriting patient health information. Convolutional neural networks (CNNs) are used for automatically determining echo views - e.g. apical 4-chamber. View-classified are then routed to one of 4 CNN models for image segmentation. We use the corresponding segmented images to compute standard measurements of cardiac structure, averaging across multiple cardiac cycles within a video and multiple videos for a given view. We also use segmentation to derive two indices of cardiac function: left ventricular ejection fraction and global longitudinal strain (GLS).  Finally, we use phased pairs of PLAX and A4c images (corresponding to end-systole and end-diastole) to detect two diseases characterized by abnormal cardiac thickening: hypertrophic cardiomyopathy (HCM) and cardiac amyloidosis.}
\label{fig:fig1}
\end{figure}

Our primary goal was to develop an analytic pipeline for automated analysis of echocardiograms that required no user intervention and thus could be deployed on a high-performance computing cluster or web application. We divided our approach into 6 steps (Figure \ref{fig:fig1}).  Preprocessing entailed automated downloading of echo studies in DICOM format, separating videos from still images, extracting metadata (such as frame rate, heart rate), converting them into numerical arrays for matrix computations, and de-identifying images by overwriting patient health information. We next used convolutional neural networks (described below) for automatically determining echo views.  Based on the identified views, videos were routed to specific segmentation models (parasternal long axis, parasternal short axis, apical-2 and apical 4-chamber) and the output was used to derive chamber measurements, including lengths, areas, volumes and estimates of mass.  Next, we generated two commonly used automated measures of left ventricular function:  ejection fraction and longitudinal strain. Finally, we phased parasternal long-axis videos to identify images at the end of cardiac systole and diastole, and used the resulting image pairs to detect two diseases:  hypertrophic cardiomyopathy and cardiac amyloidosis.  Below we elaborate on these steps, both providing technical details of performance as well as clinical applications.

\subsection{Convolutional Neural Networks (``Deep Learning'') for View Identification}
Typical echo studies consist of up to 70 separate videos representing multiple different viewpoints. For example, several different views are taken with the transducer placed beside the sternum (e.g. parasternal long axis and short axis views), at the cardiac apex (apical views), or below the xiphoid process (subcostal views). Furthermore, with rotation and adjustment of the zoom level of the ultrasound probe, sonographers actively focus on substructures within an image, thus creating many variations of these views. Unfortunately, none of these views are labeled explicitly. Thus, the first learning step involves teaching the machine to recognize individual echo views. 

For an initial model, we manually labeled six different views: apical 2-, 3-, and 4-chamber (A2c, A3c, and A4c), parasternal long axis (PLAX), parasternal short axis at the level of the papillary muscles (PSAX), the inferior vena cava (IVC) and labeled all as ``others''. We next used a multi-layer convolutional neural network, an algorithm commonly used for ``deep learning'', to distinguish between the different views.

Deep learning is a form of machine learning devised to mimic the way the visual system works \cite{lecun}. The ``deep'' adjective refers to multiple layers of ``neurons'', processing nodes tuned to recognize features within an image (or other complex input). The lower layers typically recognize simple features such as edges. The neurons in subsequent layers recognize combinations of simple features and thus each layer provides increasing levels of abstraction. The features in the top layer are typically used in a multiclass logistic regression model, which provides a final probabilistic output for classification.

\begin{figure*}[]
  \centering
  \includegraphics[width=1.0\linewidth]{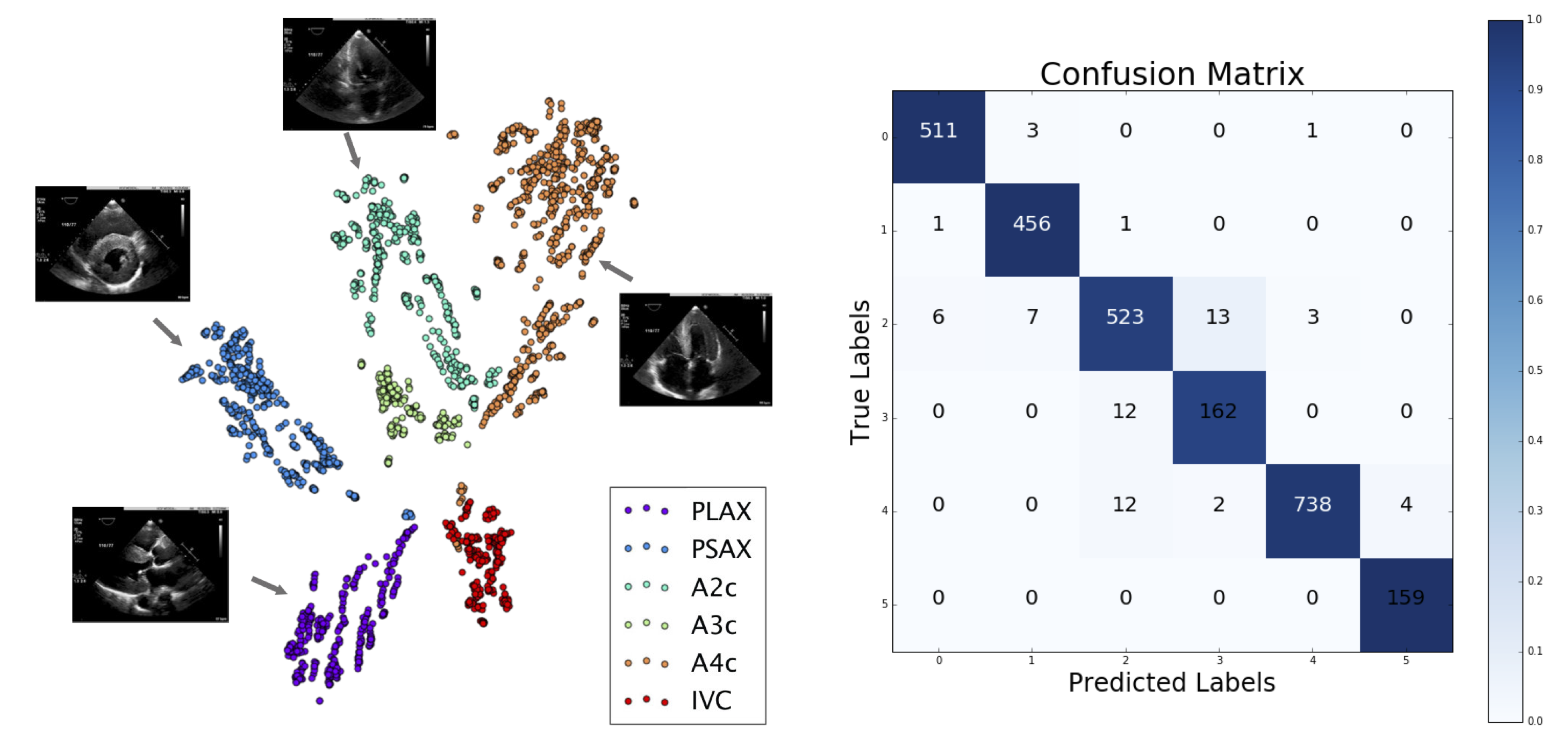}\\
  \caption[]{Convolutional neural networks successfully discriminate echo views. (Left) tSNE visualization of view classification. tSNE is an algorithm used to visualize high-dimensional data in lower dimensions. It depicts the successful grouping of test images corresponding to 6 different echocardiographic views. (Right) Confusion matrix demonstrating successful and unsuccessful view classifications within test data set. Numbers along the diagonal represent successful classifications while off-diagonal entries are misclassifications. Views are numbered as follows: 0) PLAX; 1) PSAX; 2) A2c; 3) A3c; 4) A4c; 5) IVC.}
\label{fig:fig2}
\end{figure*}

We trained a 13-layer network and found an extremely high level of accuracy for view classification as judged by cross-validation (Figure \ref{fig:fig2}, e.g. 99$\%$ for parasternal long axis). 
Clustering of the top layer features by t-Distributed Stochastic Neighbor Embedding (tSNE) \cite{maaten}, a useful algorithm for visualizing high-dimensional data, revealed clear separation of the different classes, with intuitive closer groupings of some pairs (e.g. A2c and A3c). We next trained a broader (22-class) network to enable detection of whether certain chambers are only partially visualized, as this would be essential for accurate quantification of cardiac structure and function.  For example, identifying A2c views where the left atrium is partially missing would enable excluding these videos when quantifying left atrial volumes.  For subsequent steps, we focused on PLAX, PSAX, A2c, and A4c views as these would be used to derive measures of cardiac structure and function and to develop detection models for specific diseases.

\subsection{CNNs for Image Segmentation}
Image segmentation involves identifying the location of objects of interest within an image. For example, one could identify the faces of people in a surveillance camera video or the location of other automobiles on the road in front of a self-driving car. Given that image segmentation represents a critical component of computer vision and robotics, computer scientists have developed multiple different algorithms to carry out this task.

 We initially used active appearance models for this task \cite{cootes}. However, we found that deriving a two-step approach consisting of first deriving a bounding box around the chamber of interest was error-prone.  We thus used an approach relying exclusively on CNN-based methods \cite{ronneberger} and trained separate models for PLAX, PSAX, A2c, and A4c views, which each localized 3-6 structures (Table \ref{tab:table1}). For example, for A4c, we segmented the blood pools for both the right and left atria and ventricles, the outer myocardial boundary of the left ventricle and the epicardial boundary of the whole heart.  We found very good performance for our models, with IoU values ranging from 73 to 92 for all structures of interest (the outer boundary of the entire heart was an outlier).
 \begin{table*}[] 
   \small 
   \centering 
   \begin{tabular}{| c | c | c | c |} 
   \hline
   \textbf{View} & \textbf{Number of Images Tested} & \textbf{Segmented Area} & 	\textbf{IOU Accuracy} \\ 
   \hline
   \multirow{1}{*}{Apical 2-Chamber} & \multirow{1}{*}{200} & Left atrium blood pool & 90.6 \\
                                   (A2C) & & Left ventricle blood pool & 88.1\\
                                         & & Left ventricle muscle & 72.7\\
                                         & & Outer cardiac boundary & 55.3\\
	\hline
   \multirow{1}{*}{Apical 4-Chamber} & \multirow{1}{*}{177} & Left atrium blood pool & 91.7 \\
                                   (A4c) & & Left ventricle blood pool & 89.2\\
                                         & & Left ventricle muscle & 74.0\\
                                         & & Right atrium blood pool & 81.2\\
                                         & & Right ventricle blood pool & 88.0\\
                                         & & Outer cardiac boundary & 74.0\\
	\hline
   \multirow{1}{*}{Parasternal long axis} & \multirow{1}{*}{104} & Left atrium blood pool & 83.1 \\
                                  (PLAX) & & Right ventricle blood pool & 85\\
                                         & & Aortic root & 84.7\\
                                         & & Outer cardiac boundary & 86.8\\
                                         & & Anterior septum & 76.3\\
                                         & & Posterior wall & 72.7\\
	\hline
   \multirow{1}{*}{Parasternal short axis} & \multirow{1}{*}{76} & Left ventricle blood pool & 91.9 \\
                                  (PSAX) & & Left ventricle muscle & 79.0\\
                                         & & Right ventricle blood pool & 78.6\\\hline
   \end{tabular}
   \caption[]{The U-Net algorithm trains CNN models to segment echocardiographic images.  Final column displays accuracy, as determined by cross-validation, of segmentation of specific structures within images from 4 different echocardiographic views. Segmented regions are depicted in Figure \ref{fig:fig3}.}
\label{tab:table1}
\end{table*}
 \begin{figure}[]
  \centering
  \includegraphics[width=1.0\linewidth]{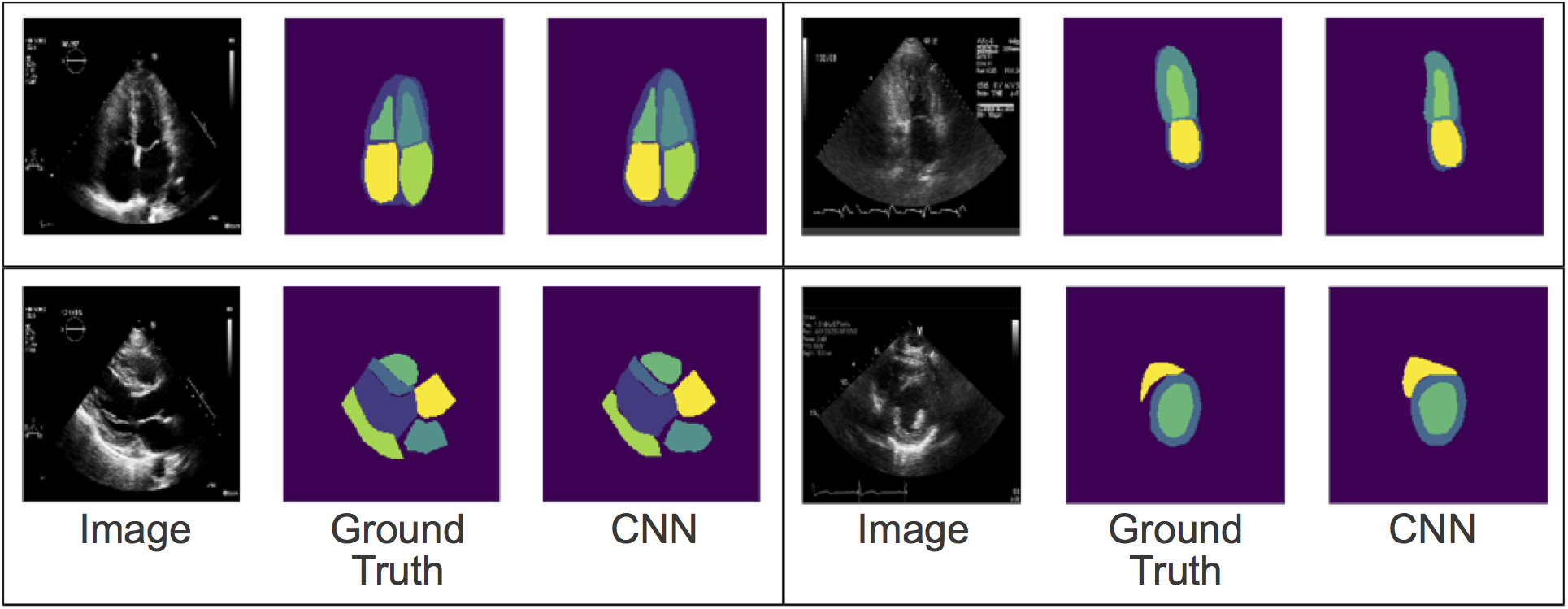}\\
  \caption[]{Convolutional neural networks successfully segment cardiac chambers. We used the U-net algorithm to derive segmentation models for 4 views: A4c (top left), A2c (top right), PLAX (bottom left) and PSAX at the level of the papillary muscle (bottom right).  For each view, the trio of images, from left to right, corresponds to the original image, the manually traced image used in training (Ground Truth) and the automated segmented image (determined as part of the cross-validation process).}
\label{fig:fig3}
\end{figure}

\subsection{Chamber Structure and Function Quantification}
As an independent ``real-world'' confirmation of segmentation, we derived commonly used measures of cardiac structure and compared our results to thousands of measurements derived from the University of California, San Francisco (UCSF) echocardiography laboratory, which uses a variety of vendor-derived software packages.  We downloaded $>$4000 studies and performed view classification and segmentation, deriving measurements according to standard guidelines \cite{lang}. For most studies, we used 6 to 8 videos for each measurement and derived a robust aggregate measure across all the studies, averaging across multiple cardiac cycles.

\begin{figure}[]
  \centering
  \includegraphics[width=1.0\linewidth]{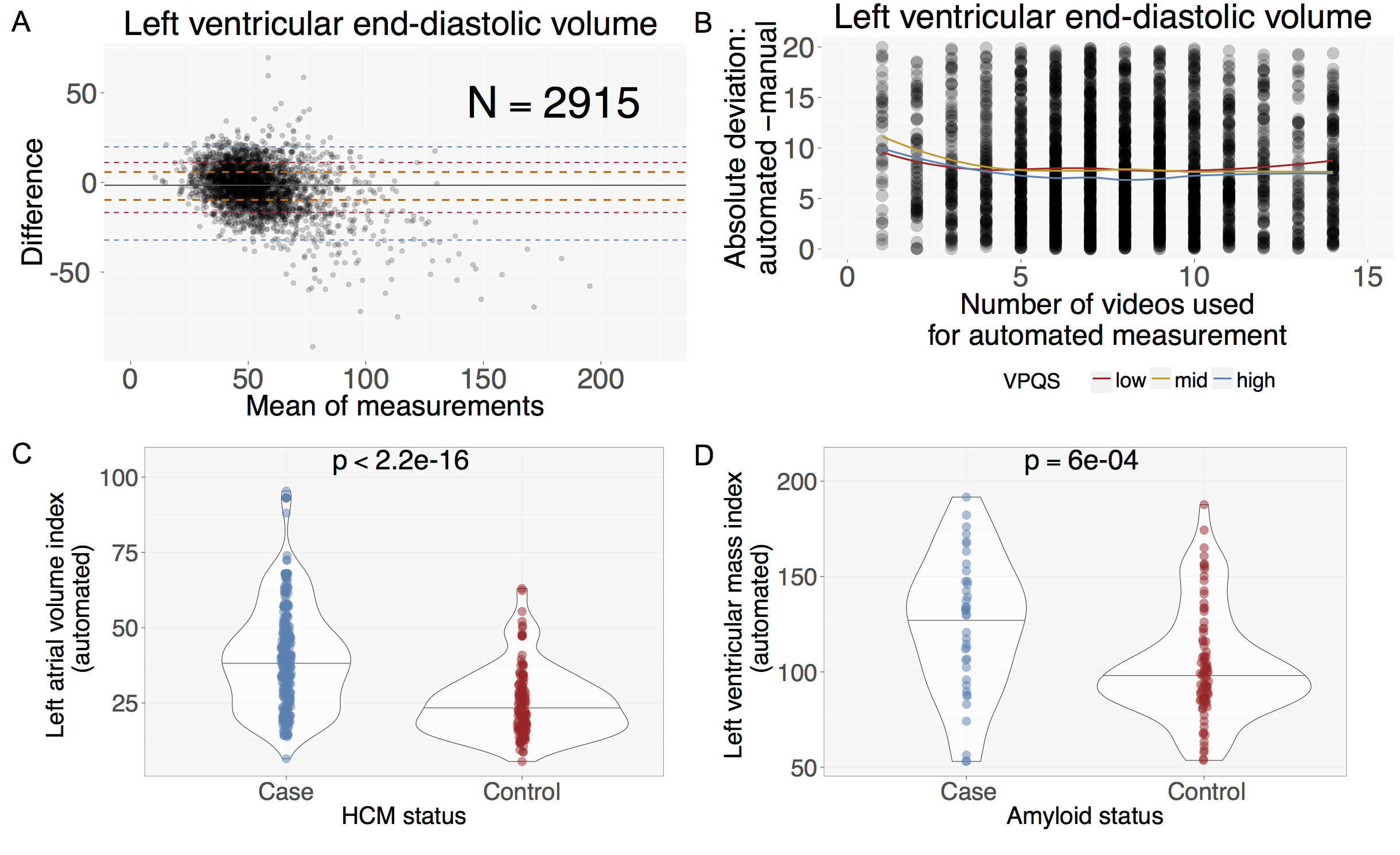}\\
  \caption[]{Automated segmentation results in accurate cardiac structure measurements in ``real-world'' conditions. (A)  Bland-Altman plot comparing automated and manual (derived during standard clinical workflow) measurements for indexed left ventricular end diastolic volume (LVEDVI) from 2915 echo studies.  Orange, red, and blue dashed lines delineate the central 50$\%$, 75$\%$ and 95$\%$ of patients, as judged by difference between automated and manual measurements.  The solid gray line indicates the median. (B) Scatter plot relating the absolute deviation between automated-manual measurements for LVEDVI and the number of videos used in the estimation. A separate loess fit is made for each of 3 tertiles of ``study quality score'' as judged by the average of the probability of view assignment across the study made by the CNN in Figure \ref{fig:fig2}. All measurements made with greater than 15 studies were binned together.   (C) Automated measurements reveal a difference in left atrial volumes between HCM patients and matched controls and (D) a difference in left ventricular mass between cardiac amyloidosis patients and matched controls.
}\label{fig:fig4}
\end{figure}

We compared our results with values derived from the routine clinical workflow (not all measurements were recorded for every study) and found excellent agreement for body-surface area indexed left atrial volume (LAVOLI) and three left ventricular measures: (indexed) left ventricular mass (LVMI), left ventricular end diastolic volumes(LVEDVI) and left ventricular end systolic volume (LVESVI), (Table. \ref{tab:table2}, Fig. \ref{fig:fig4}A). 

As an independent measure of performance, we assessed how well each method (i.e. automated vs. manual) could identify associations between different metrics. For example, it is known that the left atrium enlarges in patients with increased ventricular volume and mass - presumably reflecting increased ventricular pressures transmitted to the atria.  We found a stronger association by automated compared to manual estimation for LAVOLI vs. LVEDVI [$\rho$ = 0.50 (automated vs. automated) vs. 0.42 (manual vs. manual), N = 1366] and  LAVOLI vs. LVESVI [$\rho$ = 0.47 (automated vs. automated) vs. 0.38 (manual vs. manual), N = 1366], though a slightly weaker association for LAVOLI vs. LVMI [$\rho$ = 0.47 (automated vs. automated) vs. 0.50 (manual vs. manual), N = 825].  We also found a slightly stronger inverse relationship between automated LAVOLI and left ventricular ejection fraction [$\rho$ = -0.18 (automated vs. automated) vs. -0.15 (manual vs. manual), N = 1367], which is a measure of function. We describe a non-parametric statistical test in a later section to assess the likelihood that these differences in measures of internal consistency arise by chance.

Our approach relies on aggregation across multiple cardiac cycles within a video and across multiple videos of the same cardiac chamber. We explored to what extent each additional study contributes to the agreement between automated and measured values by fitting a linear regression model to the absolute deviation. We also generated a predictor for ``study quality'' based on the confidence with which views could be classified by the CNN described above. Specifically, we took a median of the probability of the assigned class for all videos in the study, generating a value between 0 and 1 and termed it a ``view probability quality score'' or VPQS.  We found that for LVEDVI, each 0.1 increase in VPQS reduced the absolute deviation by 2.0 mL/kg/m$^2$ (p=8x10$^{-10}$) and that each additional study used (up to 15) modestly reduced the absolute deviation by 0.02 mL/kg/m$^2$ (p=0.02) (Figure \ref{fig:fig4}B).

Changes in cardiac structure can also be useful in risk models for various diseases. For example, left atrial enlargement is a predictor for sudden cardiac death \cite{OMahony:2013ec} in hypertrophic cardiomyopathy (HCM), a disease of cardiac thickening, and one could envisage an automated standardized system to compute such measures for HCM patients. As expected, we found clear differences in left atrial volumes between HCM cases and matched controls using our automated measurements (40 vs. 25 mL/kg/m$^2$, p$<$2.2x10$^{-16}$, Figure \ref{fig:fig4}C).  Similarly, for cardiac amyloidosis, a condition described below, we found markedly increased left ventricular mass (125 vs 103 mg/kg/m$^2$, p=0.0006, Figure \ref{fig:fig4}D).
 \begin{table*}[] 
   \small 
   \centering 
   \begin{tabular}{| c | c | c | c | c | c |} 
   \hline
   \multirow{2}{*}{Metric (units described in table legend)} &
      \multirow{2}{*}{\parbox{4cm}{\centering Number of Echo Studies Used for Comparison}}&
      \multirow{2}{*}{Median Value (IQR)}&
      \multicolumn{3}{|c|}{\parbox{3cm}{\centering Absolute Deviation: Automated vs. Manual Measurement}} \\
      \cline{4-6}
    &  &  & 50$\%$ & 75$\%$ & 95$\%$\\
   \hline
   Left atrial volume index & 1452 & 27.7 (22.0--36.7) & 4.7 & 8.7 & 16.9\\\hline
   Left ventricular diastolic volume index & 2915 & 51.8 (41.8--63.8) & 7.7 & 14.0 & 26.1\\\hline
	Left ventricular systolic volume index & 2910 & 18.8 (14.3--24.9) & 4.7 & 8.5 & 16.9\\\hline
   Left ventricular mass index & 1319 & 81.0 (68.2--100.2) & 11.5 & 19.8 & 43.0\\\hline
   Left ventricular ejection fraction & 3101 & 64.0 (57.9--68.6) & 5.3 & 9.9 & 19.1\\\hline
   Global longitudinal strain & 197 & 18.0 (17.0--20.0) & 1.5 & 2.6 & 5.0\\\hline
  \parbox{4cm}{\centering Global longitudinal strain (John Hopkins PKD cohort)} & 110 & 18.0 (16.0--20.0) & 1.6 & 2.8 & 5.4\\\hline
   \end{tabular}
   \caption[]{  Comparison between fully automated and manual measurements derived from 2-dimensional echocardiography.  Absolute deviations are reported as percentiles. For each metric, 50$\%$, 75$\%$, and 95$\%$ of studies have an absolute deviation between automated and manual measurements that is less than the indicated value.  Units are mL/kg/m$^2$ for left atrial and left ventricular volumes and g/kg/m$^2$ for left ventricular mass. Ejection fraction and global longitudinal strain are dimensionless. IQR = interquartile range.}
\label{tab:table2}
\end{table*}

\subsection{Assessing Cardiac Function by Ejection Fraction and Global Longitudinal Strain}
\begin{figure}[]
  \centering
  \includegraphics[width=1.0\linewidth]{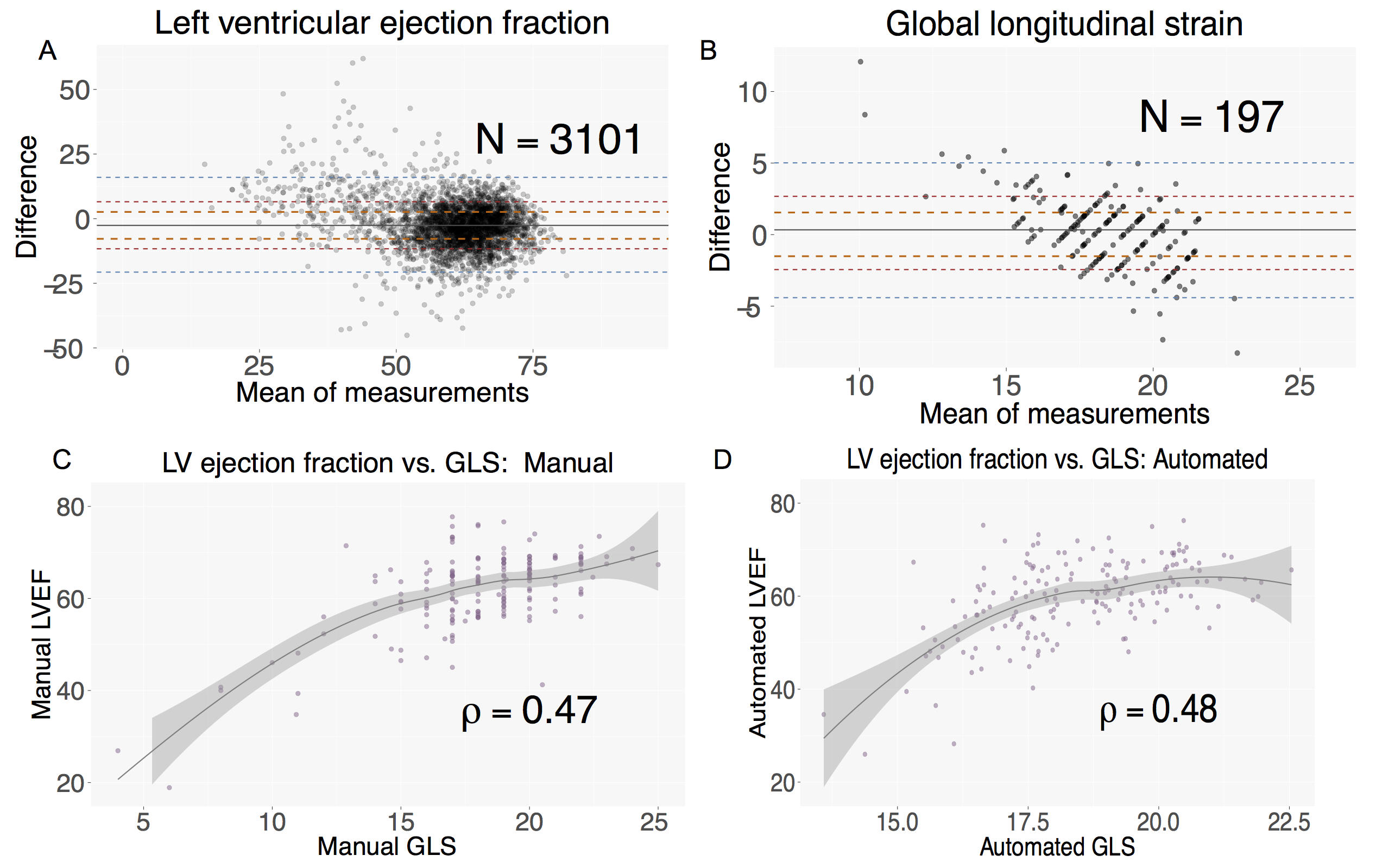}\\
  \caption[]{An automated computer vision pipeline accurately assesses cardiac function. Bland-Altman plot comparing automated and manual ejection fraction estimates for 3101 individual echo studies (A) and global longitudinal strain (GLS) for 197 echo studies (B). Delimiting lines are as in Figure \ref{fig:fig4}A. Scatter plots depicting agreement between ejection fraction and GLS for manual and automated measurements (N = 175). Spearman correlation coefficient is shown.  A loess fit with standard error is depicted.}
\label{fig:fig5}
\end{figure}

In addition to assessing the structure of the heart, 2D echocardiography provides estimates of cardiac function. The most commonly used metric, ejection fraction, can be readily computed from segmentation of the left ventricle during end diastole and end systole.  In keeping with our performance on individual left ventricular volume metrics, we found a strong if not stronger performance for ejection fraction (EF), with a MAD of 5.3$\%$ (median EF 64$\%$, N = 3101, Figure \ref{fig:fig5}A).
Along with EF, longitudinal strain is an increasingly popular method to assess the longitudinal function of the heart \cite{voigt}. It is a sensitive measure of cardiac dysfunction and is tolerant of errors in mapping of the endocardial border, whereas ejection fraction estimates depend on perfect delineation of this boundary. Although commercial packages to measure strain have been available for many years, they invariably require some user intervention and thus cannot be implemented in a scalable, fully automated pipeline. Furthermore, the black-box nature of these packages has made it difficult to interpret how the measurement is made and what limitations there may be. 

\begin{figure}[h!]
  \centering
  \includegraphics[width=1.0\linewidth]{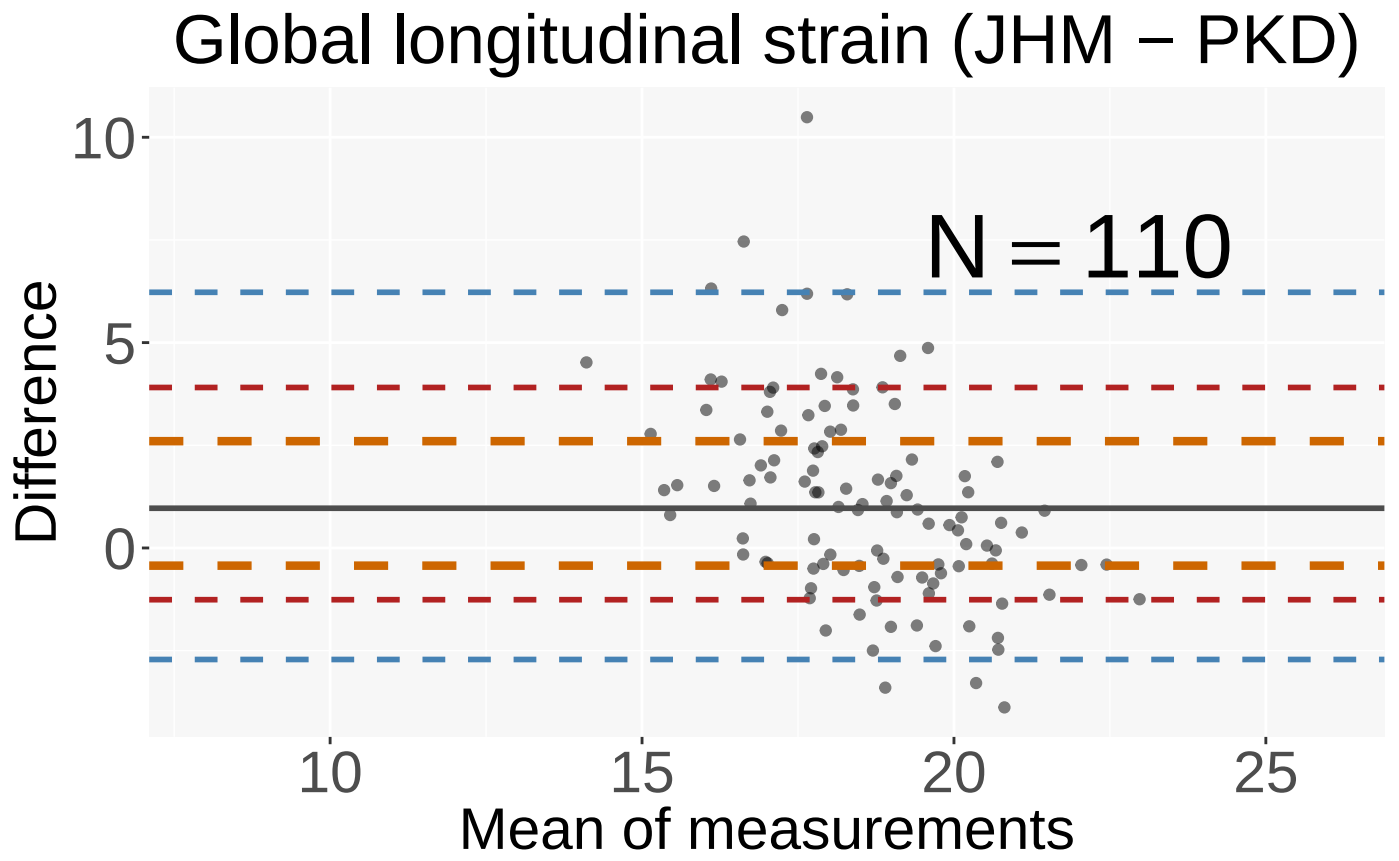}\\
  \caption{Bland-Altman plot as in Figure 5B for 110 studies from a polycystic kidney disease cohort (PKD).  Automated measurements were made blinded to manual measurements.}
\label{fig:supfig1}
\end{figure}

We wrote our own algorithm for strain estimation, adapted from a previously published approach \cite{rappaport}. We tracked echogenic particles from frame to frame to estimate velocities of particles across the length of the ventricle. Fitting this variation in particle velocity with position permitted estimates of myocardial velocity, strain rate, and strain. We compared our results to measurements based on commercial vendor packages (Figure \ref{fig:fig4}), and found excellent agreement at the patient level (MAD = 1.5$\%$, N = 197, Table \ref{tab:table2} and Figure \ref{fig:fig5}B).  Given the modest number of studies used to evaluate strain estimation compared with other metrics, we analyzed a second cohort of 110 patients from a second institution, and saw nearly identical agreement between automated and manual values (MAD = 1.6$\%$, Table \ref{tab:table2} and Figure \ref{fig:supfig1}).

As an additional evaluation of our accuracy we looked at concordance between two measures of LV function across 174 studies:  EF and GLS (Figure \ref{fig:fig5}C, \ref{fig:fig5}D). We found that the agreement between automated EF and automated GLS ($\rho$ = 0.48) was nominally better than that of manual EF vs. manual GLS values ($\rho$ = 0.47).  We also analyzed the GLS-LVEDVI association, and found stronger agreement for values generated by automation [$\rho$ = -0.19 (automated vs. automated) vs. -0.14 (manual vs. manual), N = 174] and even stronger agreement in GLS-LVESVI values for automation [$\rho$ = -0.40 (automated vs. automated) vs. -0.31 (manual vs. manual), N = 174].  Overall, across our seven metrics of internal consistency we found that our automated values were superior to those found from manual measurements (absolute average increase in Spearman coefficient = 0.05, IQR 0.03-0.06, p = 0.0198, bootstrap with 10,000 iterations).

\subsection{Mapping Patient Trajectories During Trastuzumab/Pertuzumab Treatment}
As described in the introduction, the primary motivation of this work is to facilitate early, low-cost detection of cardiac dysfunction in asymptomatic individuals to motivate initiation or intensification of therapy. Given our ability to estimate longitudinal strain accurately, we hypothesized that we should be able to use our analytic pipeline to generate quantitative patient trajectories for breast cancer patients treated with cardiotoxic agents. We identified 152 patients treated with trastuzumab or pertuzumab, antibody inhibitors of the Her2 protein, which are known to cause cardiotoxicity in a subset of patients. We downloaded 1047 echo studies from these patients and processed these through our pipeline. We generated automated plots of strain trajectories, overlaid chemotherapy usage and reported ejection fractions onto our visualization. 
 
\begin{figure}[]
  \centering
  \includegraphics[width=1.0\linewidth]{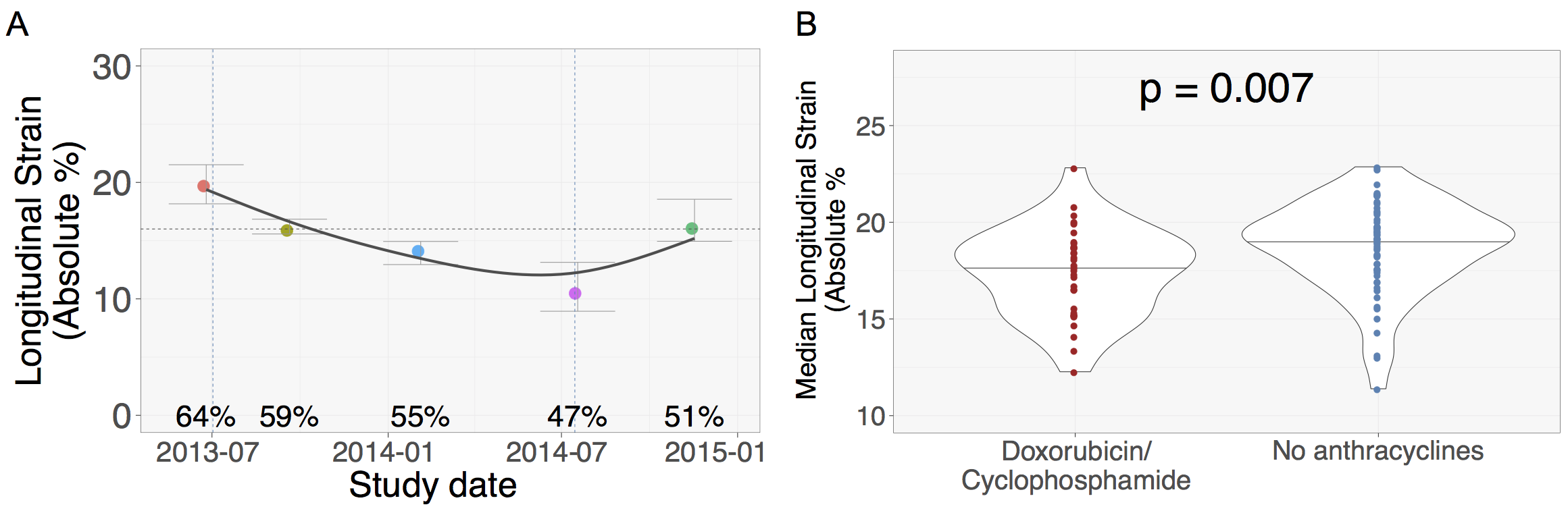}\\
  \caption[]{Automated strain measurements enable quantitative patient trajectories of breast cancer patients treated with cardiotoxic chemotherapies. Automated strain values were computed for 9421 (apical) videos of 152 breast cancer patients undergoing serial echo monitoring during chemotherapy. Individual plots were generated for each patient:  plot (A) depicts a 58 year old woman (right) receiving trastuzumab therapy only. Each colored dot represents an individual echo study. A smoothing spline was fit to the data. Ejection fractions in the published echo report are shown. Vertical blue dashed lines represent initiation and cessation of trastuzumab therapy. A horizontal dashed line at longitudinal strain of 16$\%$ indicates a commonly used threshold for abnormal strain.  (B)  Automated strain measurements confirm the more severe toxicity that occurs when combining trastuzumab/pertuzumab with anthracyclines. Violin plot showing median longitudinal strain values for patients pretreated (red) or not pretreated (blue) with neo-adjuvant doxorubicin/cyclophosphamide prior to therapy with trastuzumab (and/or pertuzumab).}
\label{fig:fig6}
\end{figure}

We observed a breadth of patient trajectories. Figure \ref{fig:fig6}A reveals an illustrative example, depicting a patient 58-year-old breast cancer patient with Type 2 diabetes mellitus and hyperlipidemia who experienced cardiac dysfunction that improved after cessation of trastuzumab, although the final strain values remains at the lower limit of normal. Such plots (with accompanying statistics) could be generated by a cloud-based interpretation system that stores prior estimates, thus allowing depiction of longitudinal trends.

 To further validate our approach, we also compared average longitudinal strain values in patients who did or did not receive doxorubicin-cyclophosphamide neo-adjuvant therapy prior to receiving trastuzumab/pertuzumab. Consistent with prior results \cite{narayan}, pretreatment with anthracyclines worsened cardiac function, as represented by lower median (19.7 vs 21.1$\%$, p = 0.01) and nadir (16.2 vs. 17.8$\%$, p = 0.02) absolute strain values (Figure \ref{fig:fig6}B).

\subsection{Models for Disease Detection}
In addition to quantifying cardiac structure and function, we sought to automate detection of rare diseases which may benefit from early recognition and specialized treatment programs. We focused on two diseases of abnormal cardiac thickening:  hypertrophic cardiomyopathy (HCM) and cardiac amyloidosis.  

HCM, which affects 0.2$\%$ of the population, is characterized by cardiomyocyte hypertrophy and disarray and myocardial fibrosis \cite{seidman}.  It can be associated with syncope, atrial and ventricular arrhythmias, heart failure, and sudden cardiac death.  Once physicians recognize HCM, they can implement preventive measures, including avoidance of high intensity exercise and implantation of a cardiac defibrillator.  Moreover, given that the first presentation of the disease can be sudden cardiac death, including in young athletes, early diagnosis can motivate physicians to screen relatives. 
 \begin{figure}[]
  \centering
  \includegraphics[width=1.0\linewidth]{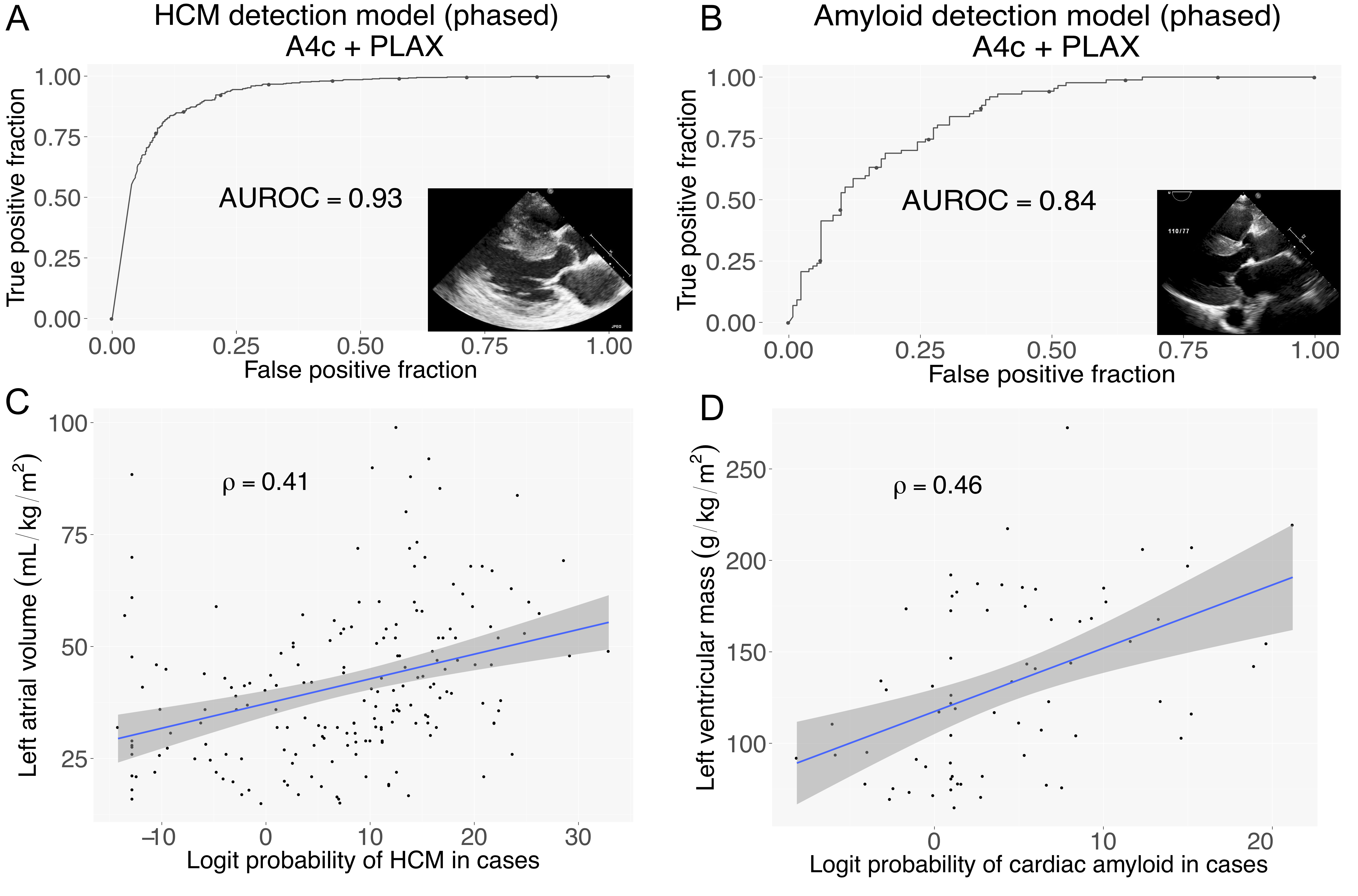}\\
  \caption[]{CNNs enable detection of abnormal myocardial diseases. Receiver operating characteristic curves for hypertrophic cardiomyopathy (A) and cardiac amyloid (B) detection.  In each case, separate CNN models were trained using hundreds of pairs of PLAX and A4c-view images for affected and unaffected individuals.  A pair consisted of one image at end-systole and one at end-diastole, where phasing was performed using estimates of the left ventricular area. Performance was assessed by cross-validation.  4-independent cross validation runs were performed and the test results averaged for each image-pair, and then a median taken across the entire study. Finally, the A4c and PLAX probabilities were averaged. (C, D) Within cases, CNN probabilities of disease were correlated with known features of the disease process (Figure \ref{fig:fig4}C,D). (C) Relationship between probability (logit-transformed) of HCM and left atrial volume with Spearman correlation coefficient indicated. (D) Relationship of probability of amyloid with left ventricular mass. Blue line indicates linear regression fit with 95\% confidence interval indicated by grey shaded area}
\label{fig:fig7}
\end{figure}
Using a cohort of HCM patients (with varying patterns of left ventricular thickening) and technically matched controls, we trained a multi-layer CNN model to detect HCM using PLAX- and A4c-view videos.  Because the heart changes appearance at different stages of the cardiac cycle, we first phased images using the results of cardiac segmentation, and selected a pair of images, one at end-diastole and one at end-systole, when the left ventricle is at its peak and minimum area, respectively.  The resulting model could detect HCM with a C-statistic (Area Under the Receiving Operating Characteristic Curve or AUROC) of 0.93. To explore possible features being recognized by the model, we plotted the (logit-transformed) probabilities of disease in cases against left atrial volume and left ventricular mass, two features associated with the disease process (Figure \ref{fig:fig4}C,\ref{fig:fig4}D). Cases with higher predicted probability of disease had larger left atria ($\rho = 0.41$, Spearman correlation coefficient) and larger left ventricular mass ($\rho = 0.38$).

We next developed a model to recognize cardiac amyloidosis, a morphologically similar yet etiologically different disease \cite{falk}.  Cardiac amyloidosis arises from the deposition of misfolded proteins within the myocardium, and can result in heart failure, bradyarrhythmias, and sudden cardiac death. Early recognition of cardiac amyloidosis can result in implementation of therapies including treatment of underlying plasma cell dyscrasias such as multiple myeloma (when the deposited proteins are immunoglobulins) or to target production of the transthyretin protein, which accumulates in other forms of the disease.  Cardiac amyloidosis can be diagnosed with confidence using cardiac biopsy or specialized imaging protocols using nuclear imaging or magnetic resonance imaging, but such specialized techniques are costly, not widely available, and thus unlikely to be deployed in many clinical settings.  Using amyloid cases and matched controls, we trained a model to cardiac amyloidosis and again found excellent performance, with a C-statistic of 0.84. Similar to HCM, we found that cases with higher predicted probability of amyloid had larger left ventricular mass ($\rho = 0.46$) but did not have increased left atrial volumes ($\rho = -0.10$).  

We have deployed our pipeline on Amazon EC2 at http://echocv.org, which features sample study outputs as well as an option to upload studies by approved users and make source code available at https://bitbucket.org/rahuldeo/echocv. 

\section{Discussion}
We achieved our primary objective, namely to construct an end-to-end automated pipeline for assessment of left ventricular structure, function, and disease detection. This pipeline is fully scalable, as evidenced by our analysis of over 4000 echo studies for this manuscript on a 10-node compute cluster all in a period of less than two weeks. Its modular nature provides multiple points for quality assessment and enables parallel improvement on multiple fronts. Although this represents the first example of a fully automated pipeline, it is important to acknowledge that many groups have made advances at several of these steps separately and insights from these efforts were invaluable for our work \cite{rappaport}\cite{bosch}\cite{hussan}\cite{narula}\cite{gao}.  The primary novelties of this work are the application of CNNs to segment echo images, the development of an empirically validated automated quality score for studies, the automation of common 2D measurements, the validation of automated values against measurements from thousands of studies, and the creation of a complete pipeline that can be deployed on the web.

Even though we have developed a working product, there is room for improvement at nearly all steps. In some cases, we anticipate that more training data will be sufficient for improved performance, although it is remarkable to note how few images ($<$200) were used to train each of our segmentation models. We are also looking at opportunities for end-to-end training to minimize cumulative errors in the pipeline, although the current modular nature of our approach enables a breadth of applications. 

Nearly all steps in our approach relies on averaging across multiple measurements and we demonstrate the utility of multiple studies in improving concordance between manual and automated measurements. Our results also suggest that health care providers that intend to use an automated computer vision pipeline for study interpretation would benefit from building more redundancy into their acquisition of echo images.  In particular, there is typically only 1 PSAX video available to compute left ventricular mass.

Encouragingly, our assessment of internal consistency - i.e. correlating different metrics such as left atrial and ventricular volumes – indicated that our measurements were modestly better than the typical clinical laboratory values. This likely arises from the ability to average across multiple measurements, a feat that would be challenging for humans forced to trace dozens of images by hand. Beyond these metrics of internal consistency, it will be important to further demonstrate utility in predicting cardiac outcomes. 

We see this work as taking a step towards augmenting clinical practice rather than replacing current approaches. Specifically, we would like to see more measurements taken when patients are asymptomatic but at risk of cardiac dysfunction, with quantitative comparisons made to prior studies to obtain personalized longitudinal trajectories. Such an approach would shift evaluation to the primary care setting, with data collected by non-experts – and the resulting initiation and tailoring of care would hopefully reduce the alarming increase in heart failure incidence that has taken place in recent decades \cite{braunwald}. A similar approach could be taken with point-of-care ultrasound at oncology infusion centers – both reducing the cost and increasing the timeliness of diagnoses of cardiotoxicity. In anticipation of such an eventuality, we deliberately avoided using any ECG information in our pipeline to accommodate analysis of data from low-cost portable handheld ultrasound devices.

Moreover, we have found that the combination of automated preprocessing and the ability to identify individual echo views using deep learning allows rapid accrual of training data for specific tasks, such as training models for the detection of mitral valve prolapse or pulmonary arterial hypertension. We are optimistic that our approach will have a broad clinical impact by 1) introducing relatively low-cost quantitative metrics into clinical practice; 2) extracting knowledge from the millions of archived echos available in echo laboratories; and 3) enabling causal insights that require systematic longitudinal tracking of patients.

\section{Materials and Methods}
\subsection{Human Subjects Research}
The institutional review board at the University of California San Francisco granted permission to download echocardiographic data as well as review relevant clinical data for this study.

\subsection{Auto-downloading of DICOM format echo studies from the Syngo client}
We realized at the onset of this project that studies stored within our echo database (Syngo, Siemens Healthcare) were in a proprietary format that could not be used for image analysis. To avoid manual download of the thousands of studies used for this project, we wrote scripts using AutoIt software (https://www.autoitscript.com/site/autoit/) to mimic human interaction with the web-based client. This enabled downloading individual studies in Digital Imaging and Communications in Medicine (DICOM) format specified by date or medical record number at a rate of approximately 1 study per 2 minutes.

\subsection{Preprocessing}
Typical echo studies consist of a combination of 80-120 still images and videos. The still images are usually used for manual measurements and thus our primary interest was in the videos. We first used the \textit{pydicom} Python library to count the number of frames within each file thus enabling separation of still images from videos. We next used the \textit{gdmconv} utility from the Grassroots DICOM Library (GDCM) to convert compressed DICOM format videos into a raw DICOM format. This allowed use of the \textit{pydicom} library for conversion of DICOM videos into numerical arrays. In doing so, we also ``blacked out'' the identifying patient information on the videos by setting the corresponding pixel intensities to minimal intensity. Numerical arrays were compressed for subsequent use. A subset of these were converted into Audio Video Interleaved (avi) format for manual segmentation.

To extract metadata corresponding to each file, we used the \textit{gdcmdump} utility from the GDCM library. We were particularly interested in the time interval between adjacent frames, heart rate, number of columns and rows, and the dimensions in physical units (i.e. centimeters) corresponding to each pixel, as these would be needed for quantitative measurements of structure and function. We removed identifying information (name, birth date) and created a compressed metadata file corresponding to each study.

\subsection{Convolutional Neural Network Based View Identification}
We based our approach on the VGG architecture by Simonyan $\&$ Zisserman \cite{simonyan}. The network takes in a fixed-sized input of grayscale images with dimensions 224x224 pixels. Each image is passed through thirteen convolution layers, five max-pool layers, and three fully connected layers. All convolutional layers consist of 3x3 filters with stride 1 and all max-pooling is applied over a 2x2 window with stride 2. The stack of convolutions is followed by two fully connected layers, each with 500 hidden units, and a final fully connected layer with size output units. The output is fed into a six-way softmax layer to represent six different echo views: parasternal long-axis (PLAX), parasternal short-axis at the papillary muscle (PSAX), apical 2-, 3-, and 4-chamber (A2c, A3c, and A4c), and inferior vena cava (IVC). The view with the highest probability was selected as the predicted view. (The final model, which focused on distinguishing occlusions as well as a broader set of views, had 22 classes). 

Additionally, each echo contains periphery information unique to different output settings on ultrasound machines used to collect the data. This periphery information details additional details collected (i.e. electrocardiogram, blood pressure, etc.). To improve generalizability across institutions, we wanted the classification of views to use ultrasound data and not metadata presented in the periphery. Because periphery information is predominantly static between frames, we tracked pixels that do not change intensity over frames and created a mask to remove such pixels. However, to account for small movement that does occur in the periphery information (i.e. ECG activity), we sampled multiple frames and removed pixels that were static for most frames.

Training data comprised of 10 random frames from each manually labeled echo video. We trained our network on approximately 40,000 pre-processed images. For stochastic optimization, we used the ADAM optimizer \cite{kingma} with an initial learning rate of 1x10$^{-5}$ and mini-batch size of 64. For regularization, we applied a weight decay of 1x10$^{-4}$ and dropout with probability 0.5 on the fully connected layers. We ran our tests for 10-20 epochs or 10-20,000 iterations, which takes ~1-2 hours on a Nvidia GTX 1080. Runtime per video was 600 ms on average.

\subsection{Convolutional Neural Networks for Image Segmentation}
Our CNN was based on the U-net architecture described by Ronneberger et al \cite{ronneberger}.  The U-net network we used accepts a 512x512 pixel fixed-sized image as input, and is composed of a contracting path and an expanding path with a total of 23 convolutional layers. The contracting path is composed of ten convolutional layers with 3x3 filters followed by a rectified linear unit and four max pool layers each using a 2x2 window with stride 2 for down-sampling. The expanding path is composed of eight convolutional layers with 3x3 filters followed by a rectified linear unit, and four 2x2 up-convolution layers. At every step in the expansion path (consisting of two convolutional layers), a concatenation with a cropped feature map from the corresponding step of the contracting path is performed to account for the loss of pixels at the border of every convolution of the contracting path. The final layer uses a 1x1 convolution to map each feature vector to the output classes. 

Separate U-net CNN networks were trained to accept as input and perform segmentation on images from PLAX, PSAX (at the level of the papillary muscle), A4c and A2c views. Training data was derived for each class of echo view via manual segmentation. We performed data augmentation techniques on training data including cropping and blacking out random areas of the echo image. Training data underwent varying degrees of cropping (or no cropping) at random x and y pixel coordinates. Similarly, circular areas of random size set at random locations in the echo image were set to 0-pixel intensity to achieve ``blackout''. This U-net architecture and data augmentation techniques enabled highly efficient training, achieving highly accurate segmentation from a relatively low number of training examples. Specifically, the PSAX segmentation U-net was trained using 72 manually segmented images, PLAX using 128 images, A4c using 168 images and A2c using 198 images. 

For stochastic optimization, we used the ADAM optimizer \cite{kingma}. Hyperparameters were optimized for each view-specific U-net, with initial learning rate set to 1x10$^{-4}$ or 1x10$^{-5}$, weight decay set to 1x10$^{-6}$, dropout set to 0.8 on the middle layers, and mini-batch size set to 5. The largest crop size and the largest blackout circle size were also tuned to each specific view with maximum crop size ranging from 40-75 pixels and maximum blackout size ranging from 30-50 pixels. We ran our tests for 150 epochs, which took approximately 2 hours for 200 training images on an Nvidia GTX 1080. In deploying the model, segmentation of each frame required 110 ms, on average.

\subsection{Automated Measurements of Cardiac Structure and Function}
We used the output of the CNN-based segmentation to compute chamber dimensions and ejection fraction. A typical echo reader typically filters through many videos to choose specific frames for measurement. They also rely on the electrocardiogram (ECG) tracing to phase the study and thus choose end-systole and end-diastole. Since our goal is to enable use of handheld echo devices without ECG capabilities, we needed to rely on segmentation to indicate the portion of the cycle. Since there are likely to be chance errors in any CNN model, we emphasized averaging as many cardiac cycles as possible, both within one video and across videos. \\

\noindent\textit{LVEDVI, LVESVI, LVEF}\\
We first used the time interval between frames and the patient heart rate to estimate the duration of the cardiac cycle. We then moved a sliding window across the video with a window length of 90$\%$ of a cardiac cycle (thus avoiding seeing end-systole or end-diastole more than once). Within a window, we selected the 90$\%$ and 10$\%$ percentile of the left ventricular volumes to serve as LV end-diastolic area and end-systolic areas, respectively.  We derived LVEDV and LVESV using the area-length formula. We also used these to compute an EF for that cycle. To enable making multiple measurements per study, we moved a sliding window across the video with a step size of half of a cardiac cycle.   We selected two additional percentile values for each metric:  one percentile applied to measurements from multiple cycles within one video, and a second across all videos in a study.  We selected the first percentile based on intuition regarding how the typical echo reader scans through images to select one for manual segmentation.  We also avoided minimum and maximum values to exclude outliers from poor quality segmentation. We selected the second percentile to minimize bias between measured and automated values, although in most cases there was relatively little difference with choice of threshold and we used the median as default. For the first cutoff (i.e. multiple measurements from one video), we used 90$\%$ percentile for LVEDVI and 50$\%$ percentile values (i.e. the median) for LVESI and LVEF.  For the second cutoff (across multiple videos in a study), we selected  median values for LVEDVI, LVESI, LVEF. \\

\noindent\textit{LAVOLI}\\
For LAVOLI, we took a similar approach, again taking the 90$\%$ percentile of the LA area for each window. If there were multiple LAVOLI measurements from one video we took the median value, and if there were multiple videos per study, we took the median of these values.  We found that erroneous LAVOLI values would arise from videos with an occluded LA. Although our view classification CNN was trained to discriminate these, some videos slipped through. We thus imposed an additional heuristic of excluding measurements from videos where LAVOLI/LVEDVI was less than 30$\%$, as we found empirically that fewer than 5$\%$ of non-occluded studies had a ratio this extreme.\\

\noindent\textit{LVMI}\\
For LVMI we again took a sliding window approach, using the 90$\%$ percentile value for the LV outer (myocardial) area and computed LVMI using the Area-Length formula \cite{lang}.  If there were multiple LVMI measurements from one video we took the median value, and if there were multiple videos per study, we took the median of these values.  

\subsection{Automated Longitudinal Strain Measurements Using Speckle Tracking}
We opted to write our own algorithm for strain computation, adapting an approach previously described by Rappaport and colleagues \cite{rappaport}. Using the results of our image segmentation, we split the left ventricle along its long axis, and output images focused on the endocardial border of the hemi-ventricle. For a given frame, we used the \textit{trackpy} Python package \cite{allan}, a particle tracking software package, to locate speckles. The \textit{trackpy} locate function allows the user to modify parameters involved in particle localization including particle diameter and minimum inter-particle separation. To track a given speckle from frame to frame, we selected a multipixel patch surrounding it and then located the best match for that patch in the next frame using the \textit{matchTemplate} function in the OpenCV package (with the TM\texttt{\textunderscore}CCOEFF\texttt{\textunderscore}NORMED statistic). Importantly, we limited the search space to that region that could be attained based on the maximum predicted velocity of the corresponding myocardial segment \cite{wilkenshoff} and excluded matches that fell below a threshold level of agreement (0.85). We then computed the displacement (in pixels) of the patch and projected the displacement onto the long axis of the ventricular segment. We fit a cubic polynomial function to estimate the variation in frame-to-frame longitudinal displacement with position along the long axis and computed its first derivative to obtain the strain rate. We next performed median smoothing and integrated the strain rate to obtain longitudinal strain. We selected the frame with the lowest (most negative) strain value across all segments to compute the global longitudinal strain, integrating both the medial and lateral portions of the ventricle.  We also computed average longitudinal strain, deriving the minimum strain value across 25-30 positions along the length of the left or right ventricle, taken separately, and then computing a median across all positions.

We noted that images with very few successfully tracked speckles gave unstable estimates of longitudinal strain and thus we adaptively lowered the threshold level of agreement to include sufficient particles for function estimation for each frame. The median number of particles that passed the original filter was stored as a measure of quality for each video’s strain estimate.

Estimation of strain typically required 1-4 minutes per video, depending on the image size and the number of frames.

\subsection{Comparison with Commercial Vendor Derived Measurements}
The UCSF echo database includes measurements for studies dating back over 15 years, although not every study includes every measurement. For downloaded studies, we extracted measurements corresponding to left ventricular and atrial volumes, ejection fraction, mass, and global longitudinal strain.  For strain, we also used echo studies collected from a second cohort of patients with Polycystic Kidney Disease seen at Johns Hopkins Medical Center. We used the identical algorithm to the one used for the UCSF data and results were generated blinded to the manual values, which were computed independently by AQ and ML using the TOMTEC (Munich, Germany) cardiac measurement software package.  

\subsection{Analysis of serial echocardiograms from Trastuzumab-and Pertuzumab-treated patients}
Patients who received trastuzumab or pertuzumab for adjuvant or metastatic disease or received a screening echocardiogram between 2011 and 2015 were identified using the UCSF pharmacy and echocardiogram databases. Patients with a transthoracic echocardiogram at baseline, early in therapy ($<$ 5 months, mean 3.0 months), and at 12 months were included in the cohort (n = 152, mean age 54.3 years, all female).  Ejection fraction values were extracted from the echocardiogram reports. Patient demographics, co-morbidities, current medications, and oncological history were obtained from chart review. 

We downloaded all available echos from each of these individuals and processed these through our entire pipeline. Plots of variation of longitudinal strain with time were generated using the \textit{ggplot2} package \cite{wickham} in R. In addition to plotting strain values, we generated a smoothing spline curve using the smooth.spline function in R.

\subsection{Downloading of echocardiograms from hypertrophic cardiomyopathy and cardiac amyloidosis patients}
We identified 225 patients who were seen at the UCSF Familial Cardiomyopathy Clinic for suspicion of hypertrophic cardiomyopathy. These patients typically had an affected family history or left ventricular hypertrophy with no clear alternative explanation. Patients had a variety of patterns of thickening including upper septal hypertrophy, concentric hypertrophy, and predominantly apical hypertrophy.  We downloaded all echos within the UCSF database corresponding to these patients and confirmed evidence of hypertrophy. We excluded bicycle, treadmill, and dobutamine stress echo studies, as these tend to include slightly modified views or image annotations that could have confounding effects on models trained for disease detection. 

Patients with cardiac amyloidosis were identified from probands seen at the UCSF Familial Cardiomyopathy Clinic and through a query of the UCSF echo database for reports including the term ``amyloid''. We identified 70 patients that had both 1) echocardiographic evidence of left ventricular hypertrophy and/or echocardiographic suspicion of cardiac amyloidosis and 2) confirmation of amyloid disease by tissue biopsy, nuclear medicine scan, cardiac MRI, or genetic testing (transthyretin variant).  We downloaded all echos within the UCSF database corresponding to these patients.  

Controls patients were also selected from the UCSF echo database.  For each HCM and amyloid case, up to 5 matched controls were selected, with matching by age, sex, year of study, ultrasound device manufacturer and model.

\subsection{CNNs to detect HCM and cardiac amyloidosis}
In addition to extracting measurements from segmentations, we also set out to develop a classifier to automate disease identification. The two diseases we targeted here are HCM and cardiac amyloidosis. Again, we based our approach on the VGG architecture by Simonyan $\&$ Zisserman \cite{simonyan} with a similar network architecture as the one used in view classification, but with 16 layers instead of 13. The stack of convolutions is followed by two fully connected layers, each with 4096 hidden units, and a final fully connected layer with 2 output units. This final layer is fed into a 2-class softmax layer to represent probabilities for HCM vs. control or amyloid vs. control.

To maintain consistency between inputs fed into the neural network, we extracted  pairs of images from each video that corresponded to end-diastole and end-systole and fed these into our neural network. Images were resized to 224x224 and consequently, our input pair had dimensions 224x224x2. To locate the end-diastole and end-systole frames in a video, we used the segmentation networks for PLAX and A4c views to extract left ventricular area values from each frame in the video. We applied a rolling median over the area values and took the frame with the 90th percentile area as the end diastole frame and the 10th percentile frame as the end systole frame.

We trained separate networks for HCM and amyloid. For stochastic optimization, we used the ADAM optimizer \cite{kingma} with an initial learning rate of 1x10$^{-5}$ and mini-batch size of 64. For regularization, we applied a weight decay of 1x10$^{-5}$ and dropout with probability 0.5 on the fully connected layers. We ran our tests for 50 epochs, which took one hour to run on an Nvidia GTX 1080. Run-time performance was approximately 600ms per video.

Accuracy was assessed using internal 5-fold cross-validation. Given that a given patient typically had multiple studies, training and test sets were defined by patient (i.e. medical record number) rather than by study. We performed four rounds of cross-validation for each view (PLAX and A4c). The score for each study $i$ was obtained by:  1) taking a median probability $p_{ij}$ across the 4-rounds of cross-validation for each video $j$; 2) taking a median of these $p_{ij}$ values for all videos in a study corresponding to a given view, resulting in $p_{PLAX_i}$ and $p_{A4c_i}$; 3) averaging the A4c and PLAX values to obtain $p_i$.

As an independent measure of interpretability of our disease detection models, we derived the Spearman correlation coefficient of $p_i$ values with left ventricular mass index and left atrial volume index values for the corresponding study, analyzing cases and controls separately.

\subsection{Statistical Analysis}
All analysis was performed using R 3.3.2. The linear regression of absolute deviation between manual and automated values of LVEDVI was computed using the standard \textit{lm} function in R. We applied a square root transform to the absolute deviation, which made the residuals approximately normal in distribution (untransformed and log-transformed values were right- and left-skewed, respectively).  To assess the chance difference between values of Spearman correlation coefficients for 7 metrics of internal consistency (LAVOLI vs. LVEF, LAVOLI vs. LVMI, LAVOLI vs. LVEDVI, LAVOLI vs. LVESVI, GLS vs LVEF, GLS vs LVEDVI, GLS vs LVESVI), we resampled with replacement (i.e. bootstrap) the input data for each comparison 10,000 times, recomputed the correlation coefficient for automated and manual values, and took the mean of the difference across all 7 metrics.  The p-value was taken as the relative frequency of observing a difference of 0 or less (i.e. manual measurements are superior) in the 10,000 iterations.

\section{Acknowledgements}
\subsection{Funding}
Work in the Deo laboratory was funded by NIH/NHLBI grant DP2 HL123228. Additional  funding  for  a  portion of echocardiographic  data  used for  validation was provided by NIH/NHLBI grant R01 HL127028 and NIH/NIDDK P30DK090868.  

\subsection{Competing Interests}
UCSF intends to submit a provisional patent based on this work.

\subsection{Author Contributions}
\noindent JZ: designed the approach to view classification, wrote code, drafted the manuscript \\
LAH: advised on computer vision approaches \\
SG: segmented images, extracted clinical data\\
LBN: performed manual strain measurements\\
MHL: performed manual strain measurements\\
GT: performed segmentation, contributed to writing the methods section\\
EF: extracted clinical data\\
MAA:  extracted clinical data\\
CJ: extracted clinical data\\
KEF: designed clinical study, edited the manuscript \\
MM: designed clinical study, extracted clinical data, edited the manuscript\\
AQ: designed clinical study, supervised MHL, edited the manuscript\\
SJS: advised on strain implementation, edited the manuscript, advised on study design, provided financial support for LBN\\
RB: advised on computer vision approaches, edited the manuscript, provided financial support for LAH and JZ\\
AE: advised on computer vision approaches, edited the manuscript, provided financial support for PA\\
PA: advised on computer vision approaches, edited the manuscript\\
RCD: designed the study, segmented images, labeled views, wrote and implemented code, performed statistical analysis, drafted the manuscript

{\small
\bibliographystyle{ieee}
\bibliography{egbib}
}

\end{document}